# EEG-Jev: A Unified Single-Pass Decision Model for EEG

**Zhifei Zhu[1,2,3,6,#]; Hongtao Zhang[1,2,3,6,#]; Yu Ren[1,2,4,6]; Wei Zhou[1,2]; Zeping Xie[1,2,5,6]; Yi He[1,2,5,6,*]**

[1]State Key Laboratory of Biomedical Imaging Science and System, Shenzhen Institutes of Advanced Technology, Chinese Academy of Sciences, Shenzhen 518055, China

[2]Paul C. Lauterbur Research Center for Biomedical Imaging, Shenzhen Institutes of Advanced Technology, Chinese Academy of Sciences, Shenzhen 518055, China

[3]Southeast University, Nanjing 211189, Jiangsu, China

[4]Harbin Institute of Technology, Harbin 150001, Heilongjiang, China

[5]Institute of Advanced Medical Devices, Shenzhen University of Advanced Technology, Shenzhen 518107, China

[6]National Innovation Center for Advanced Medical Devices, Shenzhen 518100, China

**#These authors contributed equally: Zhifei Zhu and Hongtao Zhang.**

***Corresponding author: y.he3@siat.ac.cn (Y.H.).**

**Address:** State Key Laboratory of Biomedical Imaging Science and System, Shenzhen Institutes of Advanced Technology, Chinese Academy of Sciences, 1068 Xueyuan Avenue, Shenzhen, Guangdong, China, 518055

**Abstract.** EEG analysis spans choosing alternatives, judging statements, and rating ordered attributes, requiring calibratable probabilities with explicit meaning. We present EEG-Jev, a unified single-pass decision model for EEG. Given an EEG window, a question and an answer specification, one forward pass returns a typed readout: CHOICE gives a candidate distribution, NOUL a proposition probability, and SCORE an ordered-level distribution and expectation. One checkpoint is trained jointly and evaluated once on held-out splits of 35 public sources. Core raw outputs reach 54.08% balanced accuracy on Stieger2021 motor imagery, 70.45% on Sleep-EDF-SC sleep staging, 72.11% per window at threshold 0.5 on CHB-MIT seizure detection with 0.825 AUROC, and 1.700 MAE on 0–10 ds005292 pain ratings. Prior correction and temperature scaling lower negative log-likelihood from 0.766 to 0.606 on Sleep-EDF-SC and 0.166 to 0.019 on CHB-MIT, while direct readout cuts batch-1 latency from 122.43 to 63.11 ms (1.94×) against a matched-backbone generative comparator. Interactive demonstration and aggregate results: https://neuroimglab.github.io/EEG-Jev/.

## 1 Introduction

EEG analysis serves many kinds of decisions. Some choose among alternatives, such as the sleep stage of an epoch or the movement a participant imagines. Some judge whether a statement holds, such as whether a window contains a seizure. Others rate an ordered attribute, such as the pain a participant reports. These three forms recur across clinical monitoring, brain–computer interfaces and cognitive research. In each case the user needs more than a label: a probability whose meaning is stated, that can be used as it is, and that can be calibrated.

Today these decisions are mostly served one task at a time. Pretrained EEG encoders such as LaBraM, CBraMod and REVE provide representations that transfer across datasets[1–3], but each downstream task typically still receives its own head and its own training run, and what an output means is fixed by the dataset rather than stated. EEG–language models such as NeuroLM and UniMind connect an EEG encoder to a language model so that one model covers many tasks, and they answer by generating text[4,5]. Generation is flexible, but the answer has to be parsed from the text and every answer token costs a further decoding step. A probability can be recovered by scoring the tokens of each candidate, but such scores depend on surface form and answer priors[6,7] and belong to a token sequence rather than to a stated event.

A common form for the decision itself addresses these limits. If every task is stated as a question in natural language together with an explicit specification of the admissible answers, one shared model can express tasks that differ in form. The specification also fixes what each output probability means, so that outputs can be scored and calibrated in the same way across tasks. Jev, introduced as a System One model, defines such a form: it answers typed questions about an input and returns structured answers with probabilities instead of generated text[8,9].

EEG-Jev builds a unified decision model for EEG on this form. A task is an EEG window, a question and an answer specification, and a typed readout returns the result directly: a distribution over the candidates (CHOICE), the probability that a proposition holds (NOUL), or a distribution over ordered levels with its expectation (SCORE). A frozen REVE-L encoder and a trainable connector feed the EEG to Qwen3.5-2B adapted with LoRA[1,10–12]. One set of weights is trained on all three types, and one forward pass answers a question without generating text.

The task definition is general, and our validation covers part of it: motor imagery (MI), sleep staging, seizure detection and pain rating, drawn from 35 public sources. One checkpoint is evaluated once on held-out test data for task performance and for the quality of its probabilities before and after simple post-processing, and its inference latency is measured on development windows against an autoregressive comparator built on the same backbone. Our contributions are:

1. **A typed decision form for EEG in one model.** Following the public Jev interface, EEG-Jev states choosing, judging and rating as one input form (EEG window, question, answer specification) with three typed readouts. One set of weights is trained on all three types and answers a question in one forward pass, with no text to generate or parse. In this study each application uses one readout type.
2. **Probabilities that can be scored and corrected.** Unlike EEG–language models that generate text, every output is the probability of a stated event, so proper scoring rules apply directly. A prior correction for the class-balanced training draws and one temperature per application, both fixed before the test evaluation, lower the negative log-likelihood on the sleep and seizure core sources without changing the model; decision thresholds then have to be set for the intended use.
3. **A single test evaluation across 35 sources, with latency.** One checkpoint, used without checkpoint selection, is scored once on held-out data from 35 public sources in four applications and compared with per-source baselines on the same encoder. Across the 35 sources, point estimates exceed the frozen-feature baseline on 26 sources and the fine-tuned baseline on 19; on the four core sources, the shared model matches the fine-tuned baseline on sleep staging and both baselines on pain rating, while trailing on four-class motor imagery and seizure AUROC. On a four-source development timing panel, direct readout answers a single request about 1.94 times as fast as an autoregressive comparator on the same backbone (4–6% faster at batch 32).

## 2 Related work

**EEG encoders and EEG–language models.** Pretrained EEG encoders include BIOT, LaBraM, EEGPT, CBraMod and REVE[1–3,13,14]. REVE encodes each signal patch together with the position of its electrode and its time, so one encoder accepts different electrode layouts and window lengths. We use its publicly released large checkpoint, REVE-L[10], frozen, for this property and for its public weights, not as the strongest encoder for every task: in NeuralBench-EEG the REVE entry, the base model fine-tuned end to end, is the best listed model on the core motor-imagery task, fourth on sleep staging and twelfth on seizure detection[15,16]. Among EEG–language models, NeuroLM and UniMind are instruction-tuned on several datasets and produce the answer as generated text[4,5]; the UniMind authors note that the outputs carry no probability scores. Other systems generate explanations or clinical notes[17,18] or align EEG representations with label-text embeddings or prototypes[19–21]. The Choice readout of EEG-Jev is close in form to these alignment methods, since it matches a projected state against text-embedding keys of the candidates. EEG-Jev adds proposition probabilities and rating distributions under the same task definition, trains one set of weights for the three types and evaluates all outputs as probabilities.

**Typed decisions and probability quality.** The three answer types and their output contracts follow the public interface of Jev[8,9]. The architecture and training procedure of Jev have not been published, so the model, the readouts and the training described here are our own; EEG-Jev is an independent research implementation and is not affiliated with TypeSafe. TypeSafe states that Jev is trained with a method it calls reinforcement learning for calibrated decisions (RLCD)[8,9], for which an independent open implementation has been proposed[22]; EEG-Jev uses supervised objectives only. Because candidate scores of language models depend on surface form and answer priors[6,7], we evaluate with proper scoring rules[23–25] and apply two standard corrections outside the model, prior-shift correction[26] (prior correction below) and temperature scaling[27].

## 3 Method

### 3.1 Unified task definition

We define a task without reference to any application. A task instance is a triple $(x, q, a)$: an EEG window $x$ with its electrode positions, a question $q$ in natural language, and an answer specification $a$. The specification states the answer type $\tau(a) \in \{\text{Choice}, \text{Noul}, \text{Score}\}$ and lists the admissible answers. Any task whose answer is a set of candidates, a proposition or an ordered scale fits this form, and the candidates, the proposition and the level texts are supplied with each question. Every question selects exactly one type, and the model returns a distribution over its admissible answers,

$$p_\theta(y \mid x, q, a),\ y \in \begin{cases} \{1, \ldots, K\}, & \tau = \text{Choice}, a = (a_1, \ldots, a_K) \text{ candidate texts}, \\ \{\text{no}, \text{yes}\}, & \tau = \text{Noul}, a = \phi \text{ a proposition}, \\ \{0, \ldots, J\}, & \tau = \text{Score}, a = (t_0, \ldots, t_J) \text{ levels, values } v_0 < \cdots < v_J. \end{cases} \quad (1)$$

The type fixes the event to which each probability refers. For CHOICE, $p_k$ is the probability that the reference answer is candidate $a_k$. For NOUL, $p(\text{yes})$ is the probability that the proposition $\phi$ holds for the window. For SCORE, $p_j$ is the probability that the attribute lies at level $j$, and the point answer is the expectation $\hat{y} = \sum_j v_j p_j$. The type also determines the training loss and the measures of probability quality, while the task metric follows the type and the application protocol rather than an ad-hoc choice per dataset.

### 3.2 Architecture and typed readouts

Figure 1 shows the path from an EEG window and a question to a typed output.

**EEG encoding and connector.** The frozen REVE-L encoder receives the preprocessed window with its electrode positions and returns one feature vector per channel and time patch, $f \in \mathbb{R}^{C \times P \times 1216}$. The features are flattened to $M = CP$ tokens without pooling. A trainable MLP connector maps each token to the width $d = 2048$ of the language model,

$$e_j = \gamma \;\; \mathrm{RMSNorm}\left(W_2 \mathrm{GELU}(W_1 f_j + b_1) + b_2\right), \tag{2}$$

where the learned scalar $\gamma$ is initialized to the root-mean-square magnitude of the text embeddings.

**Input row and backbone.** Each question forms one input row in the chat format of the language model:

$$[\text{system prompt}] \langle E \rangle e_1 \cdots e_M \langle /\mathrm{E} \rangle [\text{question } q][\text{answer specification } a][\text{assistant prefix}] \langle R \rangle.$$

Text is embedded with the frozen embedding table. The markers $\langle E \rangle$ and $\langle /\mathrm{E} \rangle$ and the readout marker $\langle R \rangle$ are learned vectors. The answer specification is written out as text: every candidate (“Candidate $i$: $a_i$.”), the proposition, or every level of the scale. The row passes once through the text model of Qwen3.5-2B[12,28], a causal language model with 24 layers of width 2048 that combines Gated DeltaNet and attention layers; its vision encoder is not used. Its base weights stay frozen, and rank-16 LoRA adapters on the linear maps of all layers are trained[11]. The readout marker is the last position of the row, so its final-layer state $u \in \mathbb{R}^d$ has seen the EEG, the question and the answer specification.

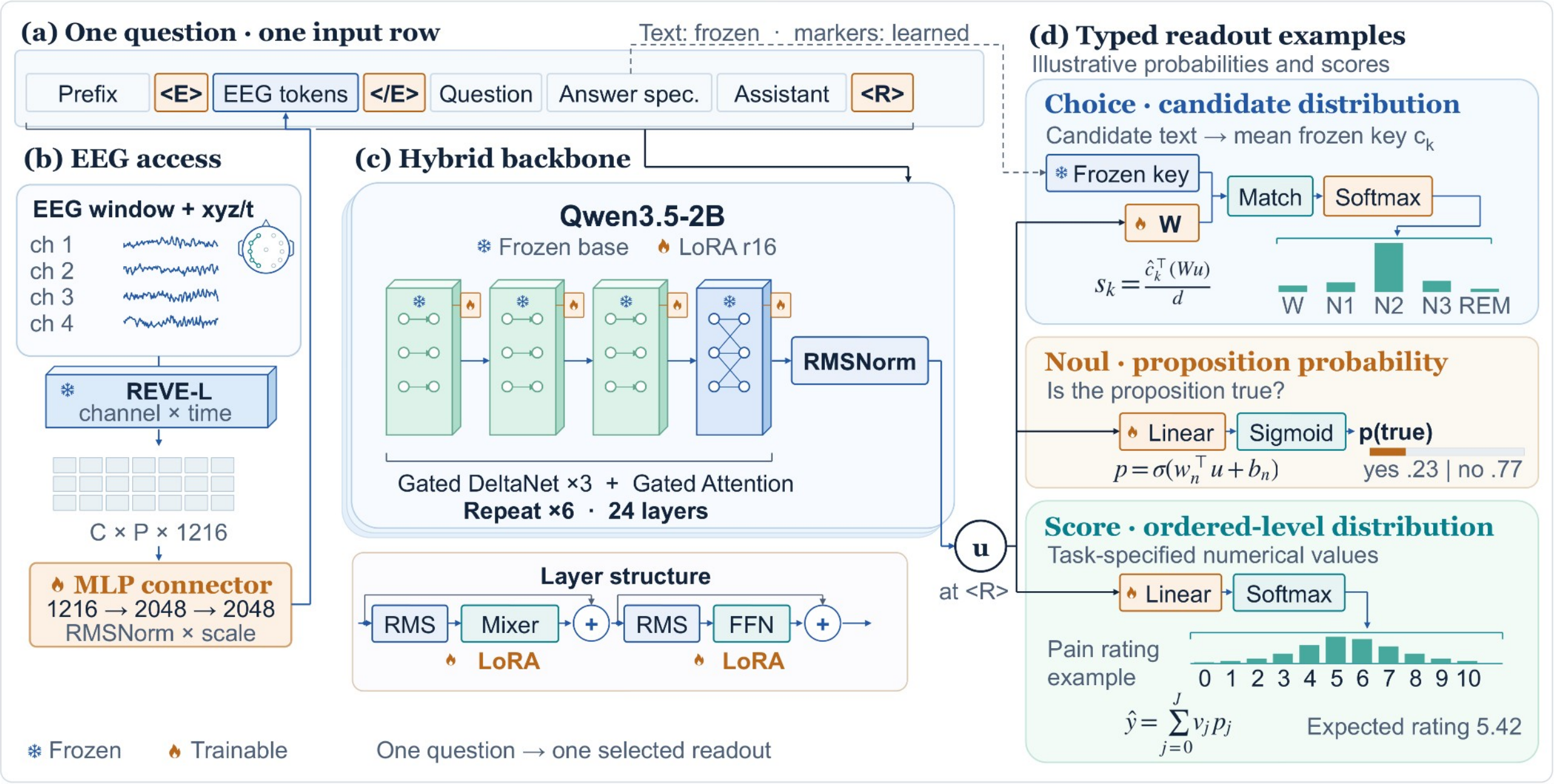

**Figure 1. EEG-Jev architecture and typed readouts.** **(a)** A prefix with the system prompt, EEG embeddings, a natural-language question, an answer specification and a readout marker $\langle R \rangle$ form one input row. **(b)** A frozen REVE-L encoder processes multichannel EEG with electrode positions and time; a trainable connector maps its features to the language-model dimension. **(c)** The Qwen3.5-2B hybrid backbone has a frozen base and trainable LoRA adapters; $u$ is the final normalized state at $\langle R \rangle$. **(d)** Each question selects one readout. CHOICE matches a trainable projection of $u$ against the candidates' keys, $\hat{c}_k = c_k / \mathrm{rms}(c_k)$, where $c_k$ is the mean of their frozen text embeddings and $d$ is the state dimension; NOUL gives the probability that a proposition holds, $p(\mathrm{yes})$, shown as p(true); SCORE gives a distribution over task-specified numerical values $v_j$ and their probability-weighted expectation. The 0–10 bars and expected rating 5.42 illustrate a pain-rating example. Probabilities, scores, EEG traces and the scalp sketch are illustrative.

**Typed readouts.** The answer type of the question selects one of three readouts; the other two are not evaluated. Together with the connector, the markers and the LoRA adapters, the readouts have 27.7 million trainable parameters; REVE-L and the 1.88 billion base weights of the language model are frozen (Appendix B).

**CHOICE.** Candidate $k$ has a key $c_k \in \mathbb{R}^d$, the mean of the frozen input embeddings of the tokens of its text $a_k$. The key does not pass through the backbone and receives no gradient. With the normalized key $\hat{c}_k = c_k / \mathrm{rms}(c_k)$ and a trainable projection $W \in \mathbb{R}^{d \times d}$, the readout matches the projected state against each key:

$$s_k = \frac{\hat{c}_k^T (Wu)}{d}, \quad p_k = \frac{\exp(s_k)}{\sum_{j=1}^{K} \exp(s_j)}. \tag{3}$$

$W$ is initialized at zero, so training starts from the uniform distribution. The readout has no class-specific output unit, so its form does not depend on the number or the wording of the candidates; we trained and evaluated it with two to five candidates.

**NOUL.** The probability that the proposition holds is $p(\mathrm{yes}) = \sigma(w_n^T u + b_n)$.

**SCORE.** A task specifies a finite set of ordered numerical support values $v_0 < \cdots < v_J$. The readout produces probabilities $p_j$ over these values and returns $\hat{y} = \sum_{j=0}^{J} v_j p_j$. Our support utilities accept 2–11 strictly increasing real values, mask unused output slots before softmax, and compute the expectation in the task's units. In the present experiments, all pain sources use the same 11-slot readout, $p = \mathrm{softmax}(W_s u + b_s)$, with $v_j = j$ for $j = 0, \ldots, 10$. Thus, 0–10 specifies the pain-rating scale used in these experiments. The three answer types follow the public Jev interface[9].

### 3.3 Shared training and typed supervision

**Question sampling and candidate reordering.** Each update uses 32 questions, and each question pairs one training window with one phrasing and the answer specification (Figure 2). Every source has a question family with five training phrasings, and one of them is drawn with equal probability for each question. For CHOICE, the candidates are permuted for each question and the target index is remapped to the new position of the annotated candidate. Target positions are balanced within each source, subject and class, so a position carries no information about the answer, and each key moves with the text of its candidate. NOUL and SCORE have nothing to reorder.

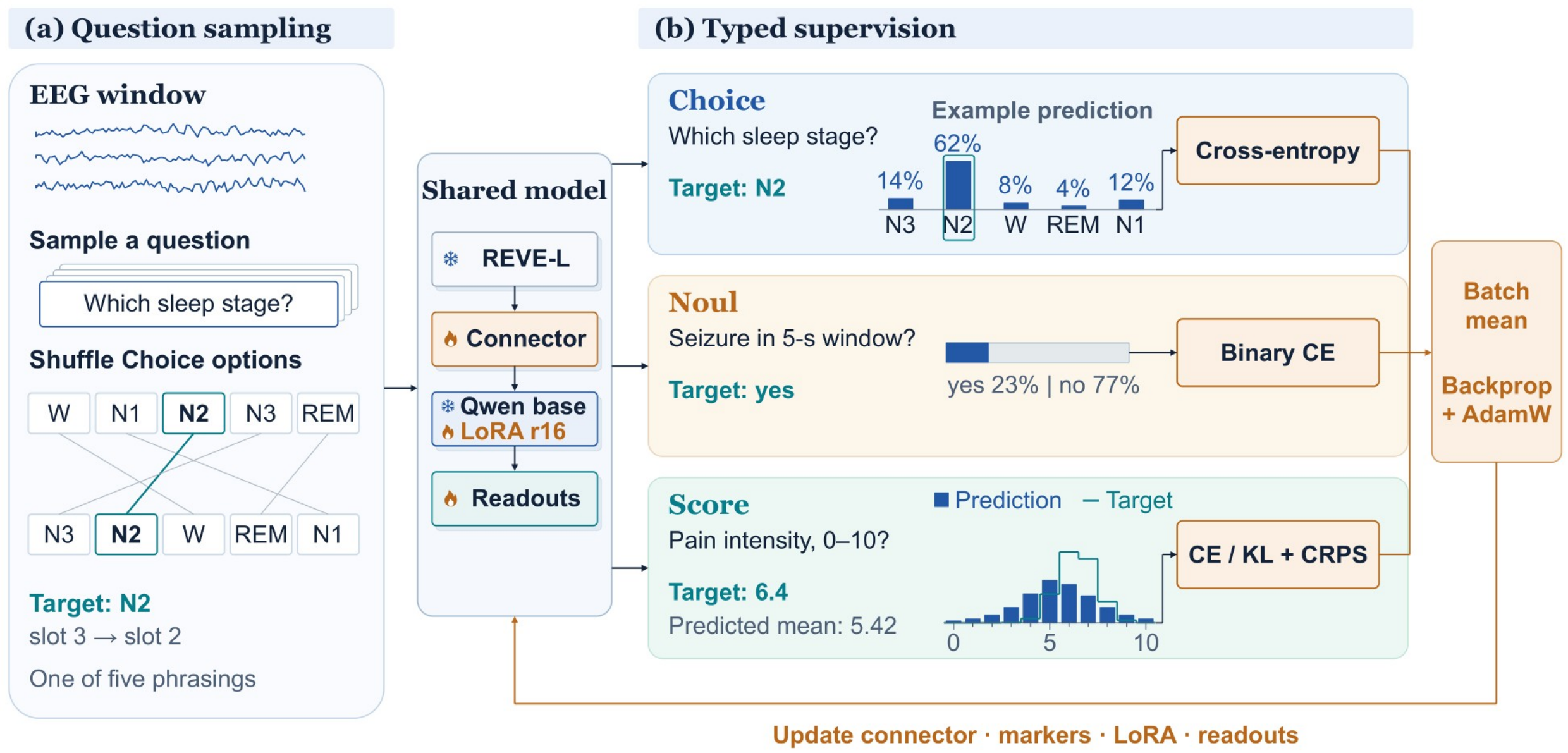

**Figure 2. Question sampling and typed supervision. (a)** Each training example pairs an EEG window with one of five question phrasings. For CHOICE the options are shuffled and the target index moves with the annotated option, here from slot 3 to slot 2. **(b)** Each training question is supervised through one type only: cross-entropy for CHOICE, binary cross-entropy for NOUL, and for SCORE a soft-target cross-entropy or a KL divergence, depending on the training stage, plus 0.2 times CRPS divided by the 10-point range. The SCORE target outline is a discrete Gaussian centred at 6.4; the prediction has mean 5.42. The batch-mean loss updates the connector, markers, LoRA and readouts; REVE-L and the backbone base stay frozen. Probabilities and EEG traces are illustrative, not model outputs.

**Typed supervision.** A question is supervised only by the objective of its own type. CHOICE uses the cross-entropy at the position of the annotated candidate, and NOUL the binary cross-entropy of $p(\text{yes})$. SCORE compares the level distribution with a soft target $r$ centred on the reported rating and adds $0.2\ \text{CRPS}/(v_J - v_0)$, where CRPS is the continuous ranked probability score. Up to update 3,600 the target $r$ splits its mass between the two levels that bracket the rating (cross-entropy); afterwards it is a discretized Gaussian with a standard deviation of one level $(\text{KL}(r \parallel p))$. The 32 per-question losses are summed and divided by 32, so every question has the same weight whatever its type and number of candidates, and one AdamW step on this mean updates all trainable parameters.

**Four stages.** One run of 13,400 updates (seed 8) has four stages (Appendix B). *Warm-up* (updates 1 to 600) trains only the connector, the markers and the readouts, with LoRA frozen, so that the EEG embeddings and the readouts first adapt to the unchanged language model. *Mixed training* (601 to 3,600) adds the LoRA adapters and draws the four applications in fixed shares. *Hard-task strengthening* (3,601 to 11,000) follows a sampling plan that was regenerated once, at update 3,600, after development panels had shown motor imagery still at chance level. The new plan sets the number of questions per source from the amount of its training data, which raises the share of motor imagery from 0.30 to 0.46 of the draws; it gives more draws to the sleep sources and stages on which the update-3,600 checkpoint was weak on a fixed development subset, and directs part of the seizure draws to training windows that this checkpoint scored poorly. Development data influenced only these sampling statistics; no development window entered a gradient. The *final stage* (11,001 to 13,400) lowers the peak learning rate tenfold and returns to shares close to those of the first two stages. The run was planned with 8,000 updates and was extended to 13,400 when the plan was regenerated. Training ended at this fixed budget: no stopping criterion was applied, the training loss was still decreasing slowly in the final stage, and the last checkpoint is used without checkpoint selection.

### 3.4 Probability post-processing

The typed outputs support post-hoc probability calibration through prior correction and temperature scaling. Training draws follow a sampling plan with class shares $\tilde{q}_k$, which can differ from the source frequencies $\pi_k$. Prior correction adjusts for this sampling-induced difference. Under the standard prior-shift model with stable class-conditional distributions, the correction takes the following form[26].

$$p_k' \propto p_k \frac{\pi_k}{\tilde{q}_k}. \tag{4}$$

Both $\tilde{q}_k$ and $\pi_k$ are counted on the training set of each source, so no development or test label is involved. SCORE has no class prior and skips this step. The prior-shift assumption holds only approximately here: the class shares of the draws changed between training stages, and from update 3,601 part of the seizure draws was selected by the model's own scores (Appendix B). We therefore treat the correction as a fixed rule to be judged by the test scores. The second step is a temperature, $p_k'' \propto (p_k')^{1/T}$, fitted per application on the development split after the prior correction[27]. It is adopted only where cross-validation over development subjects lowers the log loss.

The resulting mapping was frozen before the test evaluation: prior correction alone for motor imagery $(T=1)$; prior correction followed by a temperature for sleep staging $(T=1.078)$ and seizure detection $(T=1.252)$; a temperature alone for pain rating $(T=1.309)$. For NOUL both steps are monotone and cannot change a ranking. For CHOICE the temperature never changes a decision, but the prior correction can.

## 4 Experimental setup

The experiments ask three questions: whether one shared model reaches useful task performance across applications, whether its probabilities have the stated meaning and improve with simple post-processing, and what direct readout saves at inference time.

### 4.1 Data and evaluation scope

We train and evaluate on 35 public sources in four applications (Table 1; per source in Appendix A). Motor imagery (17 sources) is posed as CHOICE among two to four cued conditions, including rest in some sources, and sleep staging (7 sources) as CHOICE among five stages for a 30-s epoch. Each epoch is staged on its own: the model does not see the neighbouring epochs, which manual scoring rules and sequence-based automatic stagers use. Seizure detection (2 sources) is posed as NOUL with the proposition "An annotated seizure occurs at some time within this window." for a 5-s window. Pain rating (9 sources) is posed as SCORE on the 11 levels of a 0–10 scale of self-reported pain after a laser stimulus. The task is to predict this single-trial rating for subjects held out from training in that source; it is not an assessment of clinical pain.

Twenty-one sources follow the locked NeuralBench protocols with their windows, labels and partitions[15,29]: all motor-imagery sources, three sleep sources and CHB-MIT. The other 14, four sleep sources, Siena and the nine pain sources, use their own windows, labels and subject-level partitions. All sources share the signal recipe that NeuralBench uses for REVE (band-pass 0.5–99.5 Hz, 200 Hz, standardization per channel, clipping at $\pm 15$). As in the NeuralBench pipeline, the zero-phase band-pass filter and the per-channel standardization are computed over the whole recording, so the results describe offline record-level preprocessing, not a model that sees only the current window. Every source has a training, a development (validation) and a test partition. Partitions are subject-disjoint except in five motor-imagery sources whose protocols split sessions or trials of the same subjects.

**Table 1. Data and evaluation scope.** Each application uses one readout type; the core source of an application was fixed before the test evaluation and carries the main results. NB: sources that keep the windows, labels and subject splits of the NeuralBench code, release 0.4.0 (core task or full benchmark). Suppl.: supplementary sources with the same signal processing and our own windows, labels and subject splits; the nine pain sources are laser-pain studies from one research group and are not part of NeuralBench. Test: held-out subjects (patients for seizure) and windows, summed over sources; five motor-imagery sources split within subjects, and sources of one application can share participants. Per-source details: Table A1.

| Application | Sources | Readout | Answer space | Core source | Protocol scope | Test subjects / windows |
|---|---|---|---|---|---|---|
| Motor imagery | 17 | Choice | *K*=2–4 classes | Stieger2021 | 17 NB (1 core, 16 full) | 128 / 44,200 |
| Sleep staging | 7 | Choice | *K*=5 stages | Sleep-EDF-SC | 3 NB (1 core, 2 full); 4 suppl. | 291 / 293,681 |
| Seizure detection | 2 | Noul | yes / no | CHB-MIT | 1 NB (core); 1 suppl. | 9 / 155,753 |
| Pain rating | 9 | Score | 0–10, 11 levels | ds005292 | 9 suppl. | 109 / 5,080 |
| Total | 35 | | | | 21 NB; 14 suppl. | 537 / 498,714 |

One core source per application was fixed before any test result was read: the NeuralBench core tasks Stieger2021[30], the sleep-cassette recordings of Sleep-EDF (Sleep-EDF-SC)[31] and CHB-MIT[32,33], and for pain ds005292[34], selected by a rule on metadata (Appendix A). We evaluate one checkpoint (seed 8, update 13,400). It was scored once on the locked test split of every source, 498,714 windows in total, with the first training phrasing of each question family and the canonical candidate order. No test result selected a model, a threshold or a question; the one choice made after the evaluation, the displayed seizure metric, is described in Section 4.3.

### 4.2 Comparators

*B1* and *B2* are per-source baselines on the same REVE-L encoder, trained by us and scored on the same test windows as EEG-Jev. B1 is a readout on frozen features: a small MLP on each REVE-L token, mean pooling and a linear head (0.38 million trainable parameters). B2 fine-tunes the last four of the 22 encoder blocks together with the same head (71.2 million trainable parameters). Each baseline is trained and selected separately for every source with seeds 7 and 8, and we report the mean of the two seeds. *NeuralBench best* is, for each of the three NeuralBench core tasks, the best mean test result among the entries of the public NeuralBench-EEG results[15,16]. These are externally reported results of models trained or fine-tuned end to end for each task. Table 2 names the model behind each value and marks in bold the best of the three models of this study only.

### 4.3 Metrics

The main text shows one metric per application, the NeuralBench metric where a task exists: balanced accuracy (BA) for motor imagery, sleep staging and seizure detection. Pain rating has no NeuralBench task, and we report the mean absolute error (MAE) of $\hat{y}$ in rating units. Seizure BA is computed per window from the decision $p(\text{yes}) \geq 0.5$ on the raw output at the natural prevalence. The protocol named the area under the ROC curve (AUROC) as the primary seizure metric. After the evaluation, with the AUROC results known, we made BA at 0.5 the displayed metric so that all three NeuralBench applications use the NeuralBench metric; the AUROC is reported alongside. All other computed metrics are in the released result files (Appendix E).

Task results use the raw outputs. Probability quality is measured by the negative log-likelihood (NLL) and the expected calibration error (ECE, ten equal-mass bins) for CHOICE and NOUL, and for SCORE by the CRPS divided by the scale range and by the coverage of the central 80% interval. Intervals are 95% intervals of paired differences from 2,000 bootstrap resamples of the test subjects (patients for seizure) of a source. They describe the sampling of the test population only; variation between training seeds is a separate matter, which one EEG-Jev run and two baseline seeds cannot quantify. An interval that includes zero does not establish a difference between the models; no equivalence margin was set.

### 4.4 Efficiency experiment

To measure what direct readout saves (Figure 3), we compare EEG-Jev with a generative model on the same backbone that answers the same questions with short text. This comparator is initialized from the EEG-Jev checkpoint, and its LoRA adapters are then trained for 50 updates of 16 questions to produce the answer tokens followed by the end-of-sequence token, while REVE-L, the connector and the EEG markers stay unchanged. Its input row contains the same EEG embeddings, question and answer specification; the system prompt is replaced by an instruction to give one short answer, a sentence stating the answer format is appended to the question, and the readout marker is omitted. It is thus a matched-backbone autoregressive (AR) comparator adapted to short answers. It does not share the weights of EEG-Jev, and its task quality differs.

Both paths are timed from the preprocessed EEG window and its electrode positions in host memory, through the transfer to the GPU, the REVE-L encoding, the connector and the language model. This preprocessing needs the full recording (Section 4.1) and is not included in the timing. The EEG-Jev path ends when the typed probabilities and the decision are on the CPU. The AR path decodes greedily with cached states until the model emits the end-of-sequence token (at most 32 tokens) and ends when the complete text has been parsed into a valid answer on the CPU. The panel has 32 development windows from each of four sources, one per application, chosen to balance classes and subjects: Lee2019 (motor imagery; not the core source), Sleep-EDF-SC, CHB-MIT and ds005292. The two paths are timed alternately on the same windows on one GPU with 32-bit arithmetic; hardware, repetitions, synchronization and the rule for valid answers are given in Appendix D.

The *request latency* is the time of one request at batch 1; we report its mean over all requests, weighting the four sources equally. The *amortized latency per window* at batch 4 is the total time of a batch divided by 4, averaged in the same way; it is not the response time of a single request.

## 5 Results

### 5.1 Task performance across applications

Table 2 presents the four core test results. On Sleep-EDF-SC, EEG-Jev achieves 70.45% balanced accuracy, exceeding B1 by 2.29 percentage points (95% paired interval 0.46 to 3.94) and closely matching B2 (70.48%). On ds005292, its SCORE readout achieves an MAE of 1.700, compared with 1.701 for B1 and 1.717 for B2, with both paired intervals including zero. Predicting the training-set mean gives 1.771, so the three models improve on this constant reference by 0.05 to 0.07. Thus, the shared model matches the source-specific baselines on ordered rating. The pain result shows that the Score readout can be trained alongside classification and binary judgment, not that single-trial pain ratings of new subjects are predicted with clinically useful accuracy.

**Table 2. Core-source test results from raw model outputs (post-processed: Table 3).***NeuralBench best*: externally reported, the highest mean test balanced accuracy (BA) among the models listed for the task in the NeuralBench results file (commit 7aaccf3, 10 October 2026, code release 0.4.0; mean ± s.d., three seeds)[15,16]. B1 is a per-source readout on frozen REVE-L features; B2 fine-tunes the last four REVE-L blocks per source. Both are trained in this study and scored on the same test windows as EEG-Jev (mean of seeds 7 and 8). BA in percent; MAE on the 0–10 scale. Seizure BA thresholds raw $p(\text{yes})$ at 0.5 and became the displayed metric after the evaluation; on the protocol's primary metric, AUROC, EEG-Jev is below both baselines (0.825 vs 0.891 and 0.929; both paired intervals exclude zero). **Bold**: best value per row at the displayed precision among the three models of this study (EEG-Jev, B1, B2); NeuralBench best is not part of this comparison; ties bold together; bold does not indicate a significant difference. $^{\dagger}$: paired 95% interval of EEG-Jev minus this baseline excludes zero. All computed metrics: results/metrics-long.csv.

| | | | | Reported | This study | | |
|---|---|---|---|---|---|---|---|
| Application | Core source | Metric | Subjects / windows | NeuralBench best | EEG-Jev | B1 | B2 |
| Motor imagery | Stieger2021 | BA (%) ↑ | 13 / 25,340 | 68.04±0.73 (REVE-Base) | 54.08 | 61.95$^{\dagger}$ | **65.19**$^{\dagger}$ |
| Sleep staging | Sleep-EDF-SC | BA (%) ↑ | 16 / 41,656 | 72.93±0.53 (CBraMod) | 70.45 | 68.17$^{\dagger}$ | **70.48** |
| Seizure detection | CHB-MIT | BA at 0.5 (%) ↑ | 5 / 121,650 | 85.54±0.54 (BDTCN) | **72.11** | 68.81 | 68.65 |
| Pain rating | ds005292 | MAE ↓ | 26 / 779 | not covered | **1.700** | 1.701 | 1.717 |

On CHB-MIT, EEG-Jev achieves 72.11% BA at a probability threshold of 0.5, compared with 68.81% and 68.65% for B1 and B2, with both paired intervals including zero; because EEG-Jev uses 1:1 seizure draws whereas the baselines train near the natural prevalence, 0.5 corresponds to different operating points on the two raw scales. At this threshold EEG-Jev flags 6,580 of the 121,650 test windows with a sensitivity of 49.5% and a precision of 3.5%, because seizure windows are only 0.38% of the test windows. These are pooled window-level figures from five patients, not event-level detection results. Threshold-free AUROC is 0.825, compared with 0.891 and 0.929, indicating room to improve seizure ranking. On the four-class Stieger2021 task, EEG-Jev reaches 54.08% BA, below B1 (61.95%) and B2 (65.19%). B1 reads the same frozen REVE-L features as EEG-Jev, so the 7.87-point gap to B1 cannot be attributed to the frozen encoder; it arises in the connector, the language model or the joint training. EEG-Jev is below B2 on 10 of the 12 cross-subject motor-imagery sources, and on 5 of the 7 with paired intervals the interval excludes zero. These differences motivate further work on source exposure, task balance and cross-subject generalization.

Across all 35 sources, EEG-Jev exceeds B1 on 26 and B2 on 19, including 16 where it exceeds both. With AUROC, the protocol's primary seizure metric, in place of BA at 0.5, these counts are 24, 17 and 14. These counts use the displayed application metrics (BA for classification and seizure detection, MAE for pain) and compare point estimates with each baseline's two-seed mean; among the 26 sources with paired 95% intervals, intervals exclude zero in favour of EEG-Jev on 6 versus 2 against B1 and on 2 versus 6 against B2, with the remaining 18 including zero. Paired intervals and all source-level results are provided in Appendix E. Figure 4a–c shows the per-source differences with their intervals. Together, these results support the feasibility of a shared decision model across heterogeneous EEG applications. The model is ahead of the frozen-feature readout B1 on most sources and, where intervals separate the two, behind the partially fine-tuned B2 more often than ahead (6 sources against 2), with the largest deficits on cross-subject motor imagery.

### 5.2 Probability quality and post-processing

Table 3 compares the raw outputs of EEG-Jev with the frozen post-processing of Section 3.4. Figure 4d shows the NLL at each stage. Post-processing reduces probabilistic prediction error and calibration error on sleep staging and seizure detection. On Sleep-EDF-SC, NLL decreases from 0.766 to 0.606 and ECE from 0.088 to 0.044. On CHB-MIT, NLL decreases from 0.166 to 0.019 and ECE from 0.118 to 0.001. These results show that the typed outputs can be recalibrated with standard tools. Most of the gain comes from undoing the class-balanced training draws, a correction that applies to any classifier trained on resampled classes. For reference, the raw outputs of B1 and B2 (means of two seeds) have an NLL of 0.778 and 0.794 on Sleep-EDF-SC and of 0.015 and 0.014 on CHB-MIT, where a constant prediction at the test prevalence gives 0.025. On CHB-MIT the post-processed EEG-Jev output therefore improves on the constant reference but remains above both baselines, in line with its lower AUROC. On Stieger2021, the raw ECE of EEG-Jev (0.028) is lower than that of B1 (0.133) and B2 (0.183), while raw NLL is 1.093 against 1.078 and 1.130.

The stepwise results attribute most of the NLL improvement to prior correction. Seizure windows constitute half of the CHB-MIT training draws and 0.38% of its test windows. Correcting the sampling prior using training statistics brings NLL to 0.614 on Sleep-EDF-SC and 0.021 on CHB-MIT. Temperature scaling further improves probability quality; on Sleep-EDF-SC, ECE decreases from 0.057 after prior correction to 0.044. On Stieger2021, NLL decreases from 1.093 to 1.082. On ds005292, range-normalized CRPS decreases from 0.121 to 0.120 and MAE from 1.700 to 1.687.

Decision rules should account for the calibrated probability scale and the evaluation objective. On Sleep-EDF-SC, accuracy increases from 69.42% to 75.57%, while class-balanced accuracy decreases from 70.45% to 66.79%, reflecting the trade-off between frequency-weighted and class-balanced evaluation. On CHB-MIT, post-processing preserves window rankings, with AUROC at 0.825. At a threshold of 0.5, positive predictions decrease from 6,580 to 68 of 121,650 windows, sensitivity decreases from 49.5% to 13.8%, specificity reaches 99.997%, and BA decreases from 72.11% to 56.88%. Binary decisions based on calibrated probabilities therefore require a threshold chosen for a target sensitivity or false-positive cost on development data. On ds005292, the temperature moves the coverage of the central 80% interval further from its nominal value, from 82.7% to 85.6%. Probability scores, point predictions and interval coverage together characterize the post-processed outputs.

### 5.3 Inference efficiency

Figure 3 illustrates how direct typed readout produces a decision after one forward pass, while the generative path performs prefill followed by incremental decoding. With the same backbone architecture and optimized execution, EEG-Jev has a mean request latency of 63.11 ms at batch 1, compared with 122.43 ms for the adapted AR comparator, a latency ratio of 1.94. EEG-Jev is faster on all four timing sources. At batch 4, its amortized processing time is 28.68 ms per window, compared with 44.01 ms for AR, a ratio of 1.53. Figure 4e shows all batch sizes. In absolute terms, direct readout saves 59.3 ms per request at batch 1. These measurements show a latency benefit over cached autoregressive decoding for single requests and small batches, even when the generative answers average only 2.0–3.0 tokens. The comparison does not include a single-pass readout of answer-token logits or the per-source heads B1 and B2, which use no language model.

**Table 3. Probability quality of EEG-Jev on the four core test sets, raw output → output after the frozen post-processing.** The post-processing was fixed before the test evaluation: the prior correction uses class shares of the training draws and of the training set, and the temperature $T$ was fitted on development data per application (motor imagery keeps $T=1$; SCORE has no class prior). NLL: negative log-likelihood; ECE: expected calibration error with ten equal-mass bins; CRPS/range: continuous ranked probability score divided by the range of the scale; coverage: percentage of ratings inside the central 80% predictive interval (nominal 80). BA in percent. Table 2 reports the raw outputs; this table shows what the post-processing changes. Both maps are monotone for the seizure probability, so AUROC is unchanged. The post-processing moves the seizure probability to the scale of the natural prevalence, so at a threshold of 0.5 sensitivity falls from 49.5% to 13.8% and BA with it; the threshold has to be set for the intended use.

| Core source | Post-processing | Proper score | Calibration | Task metric |
|---|---|---|---|---|
| Stieger2021 | prior correction | NLL 1.093 → 1.082 | ECE 0.028 → 0.028 | BA 54.08 → 54.03 |
| Sleep-EDF-SC | prior correction, $T=1.078$ | NLL 0.766 → 0.606 | ECE 0.088 → 0.044 | BA 70.45 → 66.79 |
| CHB-MIT | prior correction, $T=1.252$ | NLL 0.166 → 0.019 | ECE 0.118 → 0.001 | AUROC 0.825; BA at 0.5 72.11 → 56.88 |
| ds005292 | temperature $(T=1.309)$ | CRPS/range 0.121 → 0.120 | 80% coverage 82.7 → 85.6 | MAE 1.700 → 1.687 |

**Figure 3. Autoregressive generation and direct typed decisions (mechanism only; measured latencies are reported in Section 5.3). (a)** EEG embeddings, question and answer specification pass the hybrid backbone in a prefill, and the language-model head yields the first token. The remaining $L-1$ tokens come from cached incremental decoding through the full backbone, which reuses the attention keys and values and the DeltaNet recurrent state; the completed sequence is parsed into an answer. $L$ is the actual number of generated tokens. **(b)** EEG-Jev takes $u$ from the final normalized state at the readout marker $\langle R \rangle$ and applies the selected CHOICE, NOUL or SCORE readout in one backbone forward pass per question. Probability bars and EEG traces are illustrative.

The timing uses fused Gated DeltaNet kernels. On the 128-window consistency panel, classification decisions agree with the reference execution, with maximum absolute changes of 0.00527 in a probability and 0.00485 in the SCORE expectation at batch 1. Task performance and calibration are evaluated with the frozen reference execution. Without the fused kernels, the reference execution gives 124.02 against 180.84 ms at batch 1 (ratio 1.46) and 34.55 against 50.02 ms per window at batch 4. The advantage shrinks with batch size, to 4–6% at batch 32 in either execution, and from batch 8 the reference execution of EEG-Jev is faster per window than the optimized one. The AR comparator receives additional short-answer adaptation and differs in task quality, so the latency comparison characterizes the two implemented decision paths under a matched backbone architecture. Appendix D provides the timing protocol, consistency checks and complete batch measurements.

## 6 Discussion

**What a unified decision model provides.** EEG-Jev brings selection, binary judgment and ordered rating into one shared model. Questions and answer specifications identify the requested decision, and typed readouts return probabilities with explicit meanings. The 35-source evaluation shows that this common interface can support four EEG applications with one set of weights. In point estimates the model exceeds the frozen-feature baseline on 26 sources and the fine-tuned baseline on 19, and it remains below both on the core motor-imagery source and on seizure AUROC. Calibration and direct inference extend its value beyond point predictions to probability-based decision systems.

**Improving performance across tasks.** The heterogeneous source sizes, class frequencies and subject distributions make training-budget allocation an important design choice. Sharing capacity across 35 sources may also introduce unequal learning progress and task interference. These are plausible contributors to the observed differences and warrant controlled experiments. We plan to study source-aware sampling, adaptive exposure based on training and validation behavior, and architectural changes that improve feature alignment and cross-subject generalization. The present results motivate these directions, while causal attribution requires matched ablations.

**Calibratable probability outputs.** Explicit probability semantics allow training-sampling priors to be corrected and temperature scaling to be applied consistently. The reductions in sleep and seizure NLL demonstrate this practical benefit. Calibration and decision thresholds should be assessed jointly for the intended use, since prior correction can improve probability quality while changing balanced accuracy and sensitivity at a fixed threshold.

**Efficiency and longer-form generation.** The matched-backbone comparison demonstrates an implementation-level latency advantage for direct typed readout at batch sizes 1 and 4. The comparator already uses short answers averaging two to three tokens. Generative systems that produce longer explanations or explicit reasoning chains require additional sequential decoding steps, which creates an opportunity for greater savings when the requested output is a typed decision. Larger backbones may add further computation. Quantifying these effects calls for comparisons with task-trained generative EEG models and reasoning-oriented systems under matched hardware, output requirements and task-quality evaluation.

**Scope of the evidence.** The results come from one training run with one seed, and the intervals describe subject sampling only. No ablation separates the contributions of the language model, of the question text or of joint training. B1 reaches 61.95% on Stieger2021 from the same frozen features with 0.38 million trainable parameters, so the present results do not show that the language model improves decoding; what it provides here is one interface for three answer types. Each application uses a single answer type, so answer type and application are confounded. Test questions use a phrasing and a candidate order that also occurred in training, so robustness to new wording and transfer to new questions are untested. All windows were preprocessed with statistics of their whole recording (Section 4.1); whether performance holds when filtering and standardization use the current window alone, as in online use on a new subject, is untested. Five motor-imagery sources share subjects between training and test by protocol, and the nine pain sources come from one research group and share some participants. The seizure results rest on five and four test patients and are pooled window-level metrics on 5-s windows: event-level sensitivity, false alarms per 24 hours and detection delay were not evaluated. With five patients the bootstrap intervals are coarse and should be read as descriptive. Sleep staging uses one epoch at a time. Development data shaped the sampling plan from update 3,600 on, when the training budget was also extended, and the displayed seizure metric was chosen after the evaluation (Section 4.3). At least seven of our sources, none of them a NeuralBench core source, are listed by NeuralBench among the pretraining data of REVE[29]; whether REVE-L saw them was not checked, and any benefit would reach all three models alike.

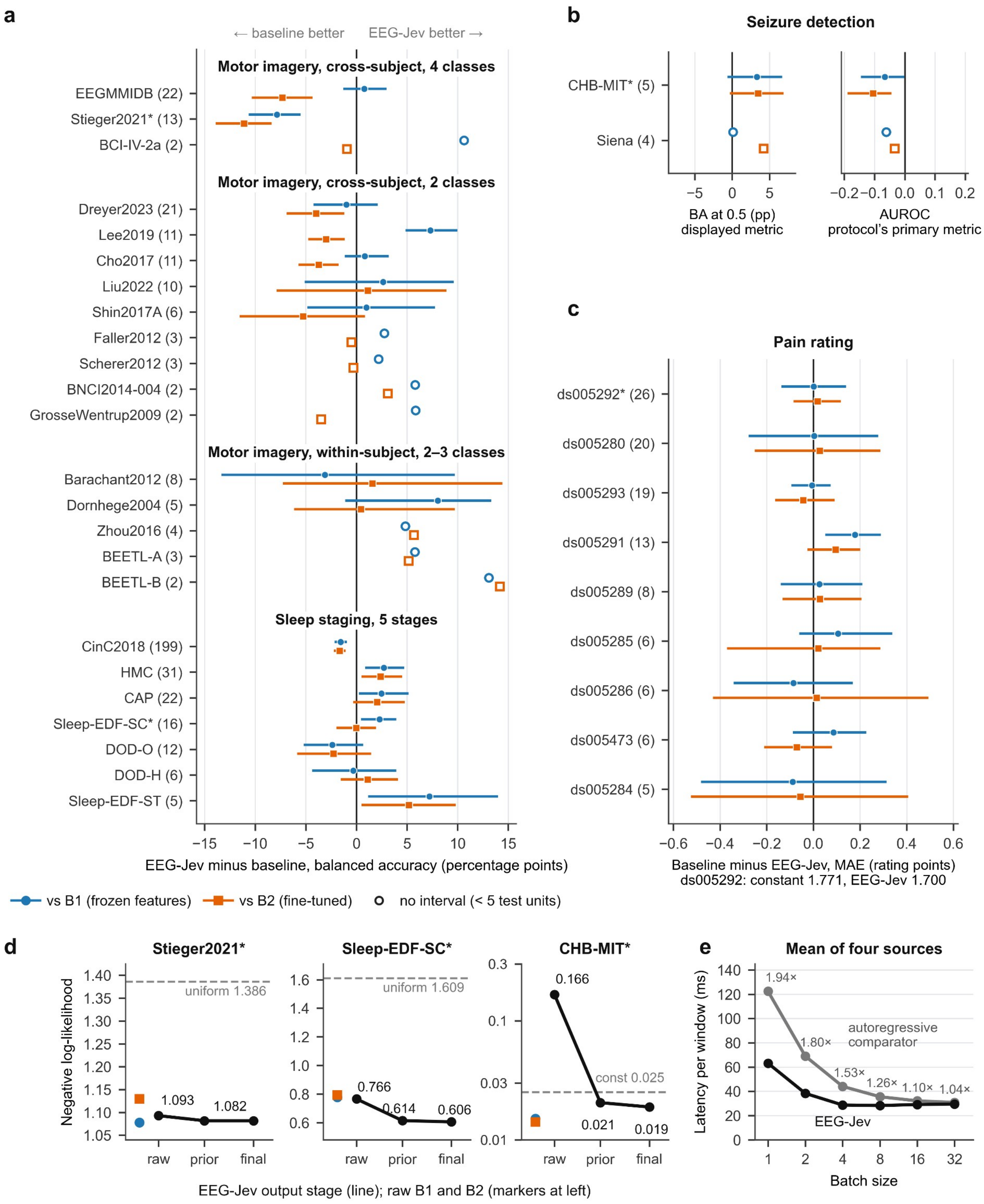

**Figure 4. Test results per source, probability scores and latency. (a)** EEG-Jev minus each per-source baseline in balanced accuracy for the 24 motor-imagery and sleep-staging sources (raw outputs; B1, frozen-feature readout; B2, fine-tuned encoder; baseline values are means of seeds 7 and 8). Bars are paired 95% intervals from 2,000 resamples of the test subjects; open

markers have no interval because the source has fewer than five independent test units. Numbers in parentheses are test units and asterisks mark core sources. **(b)** The same differences for seizure detection in the displayed metric (BA at 0.5) and in the protocol's primary metric (AUROC). **(c)** Baseline minus EEG-Jev in MAE for the nine pain sources, positive when EEG-Jev is better. **(d)** NLL of EEG-Jev on three core test sets at the raw, prior-corrected and final output stages; Stieger2021 has no temperature step, so its final stage equals the prior-corrected one. Markers at the left give the raw NLL of B1 and B2. Dashed lines give a uniform prediction over the classes (Stieger2021, Sleep-EDF-SC) or a constant prediction at the test prevalence (CHB-MIT, logarithmic axis). **(e)** Amortized latency per window against batch size on the four-source development timing panel with optimized execution, mean of four sources; numbers give the ratio of the autoregressive comparator to EEG-Jev. Data: Tables 3, A3 and A4 and results/metrics-long.csv.

**Future directions.** The first direction is scale: more datasets and applications, with a better architecture and allocation of training across sources. The second is more flexible language-conditioned decisions, including questions, candidate sets and scales that were not seen in training. The third is probability quality at training time, in particular reinforcement learning for calibrated decisions, the training method that TypeSafe names for Jev[8,22]. The fourth is deployment: distilling the typed decisions into small models that run on the device.

## 7 Conclusion

EEG-Jev unifies EEG selection, binary judgment and ordered rating in one shared decision model. Its typed readouts produce probabilities with explicit meanings in a single forward pass and support standard post-hoc calibration. Across 35 sources, it exceeds B1 on 26 and B2 on 19 in point estimates under the displayed application metrics; among the 26 sources with paired intervals, the interval excludes zero in its favour on 6 and against it on 2 versus B1, and on 2 and 6 versus B2. The clearest remaining gaps are on cross-subject motor imagery and seizure ranking. Sleep and seizure probabilities benefit from prior correction and temperature scaling. On the four-source timing panel, direct readout achieves a mean single-request latency of 63.11 ms and a batch-4 amortized processing time of 28.68 ms per window, with latency ratios of 1.94 and 1.53 relative to the matched-backbone generative comparator. Together, these findings provide a practical basis for efficient, calibratable EEG decisions and motivate further development of training allocation, generalization and model architecture.

## Declarations

**Data availability.** All 35 sources are public datasets (Appendix A). None requires an application or a signed data-use agreement; two (Shin2017A, Dornhege2004) ask users to accept terms on the download page. Dataset access information and licensing details are provided through the project website. Sources hosted on PhysioNet are cited as their pages request[35]. The project website redistributes no raw EEG recordings, derived features, per-window test predictions or model weights; aggregate results are released as machine-readable files. The accompanying demonstration replays short display excerpts of development-split recordings, with the saved model outputs for those excerpts, only for sources whose EEG data licences permit redistribution with attribution. Licensing ambiguities and third-party terms for individual sources are documented in the registry.

**Ethics and use.** This is a secondary analysis of public, de-identified EEG recordings and involved no new recruitment or data collection. The model and its results are for research only and are not intended for clinical diagnosis, monitoring or alarm.

**Acknowledgements.** We thank the investigators and data-sharing initiatives that made the EEG datasets used in this study publicly available, and the developers and maintainers of the open-source software and models on which EEG-Jev builds, including NeuralBench, REVE and Qwen.

## CRediT authorship contribution statement

**Zhifei Zhu:** Conceptualization, Methodology, Software, Validation, Formal analysis, Investigation, Visualization, Writing-original draft; **Hongtao Zhang:** Conceptualization, Methodology, Software, Validation, Writing-review & editing; **Yu Ren:** Data curation, Validation, Writing-review & editing; **Wei Zhou:** Resources, Validation, Writing-review & editing; **Zeping Xie:** Resources, Validation, Writing-review & editing; **Yi He:** Conceptualization, Supervision, Funding acquisition, Resources, Investigation, Writing-review & editing.

**Competing interests.** The authors declare no competing interests.

**Funding.** This work was supported by grants from the Strategic Priority Research Program of the Chinese Academy of Sciences (XDB0930000), the Shenzhen Science and Technology Program (JSGGKQTD20210831174329010), the Shenzhen Outstanding Talents Training Fund, the National Key Research and Development Program of China (2023YFF0714200, 2024YFC3406602), the CAS Project for Young Scientists in Basic Research (YSBR-114), the Key Laboratory for Magnetic Resonance and Multimodality Imaging of Guangdong Province (2023B1212060052), and the National Natural Science Foundation of China (No. 82505533).

## A Sources and partitions

Table A1 lists every source with its protocol, readout, partition and test scope. The pain core source ds005292 was chosen before any pain test result by a rule on metadata: among the pain sources that were in training from the first update, used the common experimental design of the series and had no recorded contact with their test data, it has the most training and development subjects. The per-trial ratings of the nine pain sources are not part of the OpenNeuro records; they were taken from the supplementary repository of the dataset authors, which carries no licence statement (see the registry).

**Table A1. The 35 sources.** Protocol: core task or full benchmark of the NeuralBench code, release 0.4.0 (NB core, NB full), or supplementary source (Suppl.). Readout with the number of options (CHOICE) or levels (SCORE). Split: subjects disjoint across partitions (cross) or shared (within). Subjects: patients for seizure; test counts and test windows are those of the held-out test results. $*$: core source. [a] input channel defect shared by all compared models; [b] ds005284 and ds005292 share participants; [c] ratings were read before the partition was built; [d] joined training at update 3,601. Reference: primary dataset paper (all nine laser-pain datasets share the omnibus data descriptor[34]); exact repository DOIs, versions and access points are recorded in the registry. Licence: short name of the data licence ([e] further statements or conditions are recorded in the registry; for the nine pain sources the EEG records are CC0, and the per-trial pain ratings come from a supplementary repository of the dataset authors that carries no licence statement); the data registry on the project website (results/data-source-registry.csv) gives the version, access point, licence text and conditions of every source.

| Source | Reference | Licence | Protocol | Readout | Split | Subjects train / dev. / test | Test windows |
|---|---|---|---|---|---|---|---|
| *Motor imagery* | | | | | | | |
| Stieger2021* | 30 | CC BY 4.0[e] | NB core | Choice, 4 | cross | 36 / 13 / 13 | 25,340 |
| EEGMMIDB | 36 | ODC-By 1.0 | NB full | Choice, 4 | cross | 63 / 21 / 22 | 1,980 |
| BCI-IV-2a | 37 | CC BY-ND 4.0 | NB full | Choice, 4 | cross | 5 / 2 / 2 | 1,152 |
| Dreyer2023 | 38 | CC BY 4.0 | NB full | Choice, 2 | cross | 52 / 14 / 21 | 5,040 |
| Lee2019 | 39 | CC0 1.0 | NB full | Choice, 2 | cross | 32 / 11 / 11 | 2,200 |
| Cho2017 | 40 | CC0 1.0 | NB full | Choice, 2 | cross | 30 / 11 / 11 | 2,200 |
| Liu2022 | 41 | CC BY 4.0 | NB full | Choice, 2 | cross | 30 / 10 / 10 | 400 |
| Shin2017A | 42 | GPL v3[e] | NB full | Choice, 2 | cross | 17 / 6 / 6 | 360 |
| Faller2012 | 43 | CC BY-NC-ND 4.0 | NB full | Choice, 2 | cross | 6 / 3 / 3 | 1,200 |
| Scherer2012 | 44, 45 | CC BY-ND 4.0 | NB full | Choice, 2 | cross | 8 / 3 / 3 | 480 |
| BNCI2014-004 | 37, 46 | CC BY-ND 4.0 | NB full | Choice, 2 | cross | 5 / 2 / 2 | 1,440 |

| Source | Reference | Licence | Protocol | Readout | Split | Subjects train / dev. / test | Test windows |
|---|---|---|---|---|---|---|---|
| GrosseWentrup2009[a] | 47 | CC BY 4.0 | NB full | Choice, 2 | cross | 6 / 2 / 2 | 600 |
| Barachant2012 | 48 | CC BY-SA 4.0 | NB full | Choice, 3 | within | 8 / 8 / 8 | 96 |
| Dornhege2004[a] | 49, 50 | see registry | NB full | Choice, 2 | within | 5 / 5 / 5 | 112 |
| Zhou2016 | 51 | CC BY 4.0 | NB full | Choice, 3 | within | 4 / 4 / 4 | 600 |
| BEETL-A | 52 | CC BY 4.0 | NB full | Choice, 3 | within | 3 / 3 / 3 | 600 |
| BEETL-B | 52 | CC BY 4.0 | NB full | Choice, 3 | within | 2 / 2 / 2 | 400 |
| *Sleep staging* | | | | | | | |
| Sleep-EDF-SC[*] | 31 | ODC-By 1.0 | NB core | Choice, 5 | cross | 46 / 16 / 16 | 41,656 |
| HMC | 53 | CC BY 4.0 | NB full | Choice, 5 | cross | 90 / 30 / 31 | 28,766 |
| CinC2018 | 54 | ODC-By 1.0 | NB full | Choice, 5 | cross | 596 / 199 / 199 | 174,033 |
| CAP[a] | 55 | ODC-By 1.0 | Suppl. | Choice, 5 | cross | 63 / 10 / 22 | 21,678 |
| DOD-H | 56 | MIT | Suppl. | Choice, 5 | cross | 15 / 2 / 6 | 5,903 |
| DOD-O | 56 | MIT | Suppl. | Choice, 5 | cross | 33 / 5 / 12 | 11,349 |
| Sleep-EDF-ST | 31 | ODC-By 1.0 | Suppl. | Choice, 5 | cross | 13 / 4 / 5 | 10,296 |
| *Seizure detection* | | | | | | | |
| CHB-MIT[*,a] | 32, 33 | ODC-By 1.0[e] | NB core | Noul | cross | 13 / 5 / 5 | 121,650 |
| Siena[a] | 57 | CC BY 4.0 | Suppl. | Noul | cross | 8 / 1 / 4 | 34,103 |
| *Pain rating* | | | | | | | |
| ds005292[*,b] | 34 | CC0 1.0[e] | Suppl. | Score, 11 | cross | 83 / 25 / 26 | 779 |
| ds005280[a] | 34 | CC0 1.0[e] | Suppl. | Score, 11 | cross | 60 / 20 / 20 | 600 |
| ds005284[b] | 34 | CC0 1.0[e] | Suppl. | Score, 11 | cross | 16 / 5 / 5 | 80 |
| ds005285 | 34 | CC0 1.0[e] | Suppl. | Score, 11 | cross | 17 / 6 / 6 | 960 |
| ds005286 | 34 | CC0 1.0[e] | Suppl. | Score, 11 | cross | 18 / 6 / 6 | 180 |
| ds005289[c] | 34 | CC0 1.0[e] | Suppl. | Score, 11 | cross | 23 / 8 / 8 | 78 |
| ds005291 | 34 | CC0 1.0[e] | Suppl. | Score, 11 | cross | 39 / 13 / 13 | 390 |
| ds005293 | 34 | CC0 1.0[e] | Suppl. | Score, 11 | cross | 57 / 19 / 19 | 1,243 |
| ds005473[d] | 34 | CC0 1.0[e] | Suppl. | Score, 11 | cross | 17 / 6 / 6 | 770 |

## B Training and configuration

**Model.** The released REVE-L[10] has 22 blocks of width 1216 and cuts each channel into patches of 200 samples with a stride of 180, which gives 3, 4, 5 and 33 patches for windows of 3, 4, 5 and 30 s. Qwen3.5-2B has 24 layers in six groups of three Gated DeltaNet layers and one attention layer. LoRA (rank 16, $\alpha=32$) is applied to all 186 linear maps of these layers. The trainable parameters are 6,690,817 in the connector, 6,144 in the three markers, 16,819,200 in LoRA and 4,218,892 in the readouts (4,194,304 of them in the CHOICE projection $W$), 27,735,053 in total; an inherited linear layer of 22,539 parameters that is never called is not counted.

**Schedule.** Table A2 gives the four stages. Training ended at the fixed budget of the regenerated plan, without a stopping criterion; the training loss was still decreasing slowly in the final stage. The optimizer is AdamW ($\beta=(0.9, 0.999)$, $\epsilon=10^{-8}$, weight decay 0.01) with gradient clipping at norm 1.0 and 32-bit arithmetic. Each update averages the losses of 32 questions, 428,800 training questions in total. The run took 10.4 hours on one NVIDIA RTX PRO 6000 Blackwell GPU.

**Table A2. Training stages of the single run (seed 8).** Peak learning rates are given for the connector, markers and readouts and for LoRA. In each stage the learning rate rises linearly over the ramp (number of updates; none in the final stage) and then follows a cosine decay to one tenth of the stage peak. Before the plan was regenerated at update 3,600, the last two stages were planned with 2,600 and 1,800 updates, 8,000 in total. Shares are the fractions of questions drawn from motor imagery, sleep staging, seizure detection and pain rating; those of the third stage result from the per-source question counts of the regenerated plan.

| Stage | Updates | Trained modules | Peak learning rate | Shares (MI/sleep/seizure/pain) | Ramp |
|---|---|---|---|---|---|
| Warm-up | 1–600 | connector, markers, readouts | $10^{-3}$ / LoRA frozen | 0.25 / 0.61 / 0.04 / 0.10 | 10 |
| Mixed training | 601–3,600 | those of warm-up, and LoRA | $3\times10^{-4}$ / $2.5\times10^{-5}$ | 0.30 / 0.56 / 0.04 / 0.10 | 100 |
| Hard-task strengthening | 3,601–11,000 | as in mixed training | $3\times10^{-4}$ / $2.5\times10^{-5}$ | 0.46 / 0.39 / 0.04 / 0.11 | 100 |
| Final stage | 11,001–13,400 | as in mixed training | $3\times10^{-5}$ / $2.5\times10^{-6}$ | 0.25 / 0.62 / 0.04 / 0.09 | 0 |

**Sampling.** Within a stage, a deterministic scheduler serves the source that is furthest below its share. In the first two stages the motor-imagery and sleep sources are weighted by the square root of their number of training subjects, classes are drawn equally, seizure draws are balanced between positive and negative windows, and pain draws cycle through all rated events. The plan regenerated at update 3,600 sets the question count of each source for the whole run from the amount of its training data (for Lee2019 and Cho2017, from their exposure in an earlier run) and subtracts what was already drawn. For sleep, a fixed development subset of 16,717 windows, scored with the update-3,600 checkpoint, sets source multipliers between 0.81 and 1.19 from the BA of each source and class shares between 0.12 and 0.35 from the per-class F1, highest for stage N1. For seizure, 30% of the positive draws come from the fifth of positive training windows with the lowest $p(\text{yes})$ under that checkpoint and 50% of the negative draws from the tenth of scored negative training windows with the highest. The ninth pain source, ds005473, enters at update 3,601, where the SCORE target also changes (Section 3.3). Per-source question counts will be released with the codebase.

**Baselines.** B1 applies an MLP $(1216 \rightarrow 256 \rightarrow 256)$ to each REVE-L token, followed by mean pooling, layer normalization and a linear head. B2 runs REVE-L on the preprocessed window and fine-tunes its last four blocks with the same head. For each source a fifth of the training subjects, at least two, is held out for model selection, and CHOICE losses are weighted by inverse class frequency. The baselines are references: their learning rate was selected from two values, and nothing else was tuned. They are not fully uniform. Seven B1 sources (Lee2019, Cho2017 and five pain sources) reuse readouts trained earlier with the same architecture, the seed-7 B2 runs come from three code paths with the same split, loss and selection rule, and ten seed-8 B2 runs ended at the limit of ten epochs.

## C Question examples

Each source belongs to one of seven question families, and each family has five training phrasings. Evaluation uses the first of them:

- Sleep staging (CHOICE): “Which sleep stage is assigned to this 30-second EEG epoch under the dataset annotation protocol?”

- Motor imagery (Choice): “Which hand movement did the participant imagine during this cued motor-imagery trial?”, with candidates such as “Candidate 1: Left-hand movement. Candidate 2: Right-hand movement.”
- Seizure detection (Noul): “Does any annotated seizure interval overlap this 5-second EEG window with positive duration?”, with “Proposition: An annotated seizure occurs at some time within this window.”
- Pain rating (Score): “What pain intensity did the participant report for this laser-stimulation trial on the 0-to-10 rating scale?”, with one line per level, “Level $k$: Reported pain intensity: $k$ out of 10.”

The AR comparator of the efficiency experiment receives the same question and answer specification. Its system prompt is replaced by “Answer the question about the EEG with one short answer only. Do not explain.”, and a sentence stating the answer format is appended to the question, for example “Answer only yes or no.” All phrasings will be released with the codebase.

**D Complete efficiency measurements**

Table A3 reports the measurements for every source and batch size under the optimized configuration. The accompanying benchmark file also contains the reference execution measurements, enabling comparison across kernel configurations and batch sizes.

**Table A3. Inference benchmark of EEG-Jev against the matched-backbone autoregressive comparator adapted to short answers (AR), same GPU, precision and kernels, 32 development windows per source.** Top: request latency at batch size 1 (96 timed requests per source and path), number of generated tokens (end-of-sequence token included) and task quality on the 32 timing windows (BA in 0–1; MAE for ds005292), which is not a test result. Bottom: amortized latency per window (total batch time divided by the number of windows) for batch sizes 1–32, EEG-Jev / AR; the last column is the equal-weight mean over the four sources. Timing starts from the preprocessed EEG in host memory and ends with the decision (EEG-Jev) or the parsed answer (AR) on the CPU. All values: results/inference-benchmark.csv.

| | Request latency at batch size 1 (ms) | | | | | | | Panel quality | |
|---|---|---|---|---|---|---|---|---|---|
| | EEG-Jev | | | AR | | | AR tokens | | |
| Source | mean | p50 | p95 | mean | p50 | p95 | mean | EEG-Jev | AR |
| Lee2019 | 61.16 | 56.30 | 77.74 | 143.30 | 144.22 | 172.03 | 3.00 | 0.781 | 0.719 |
| Sleep-EDF-SC | 59.16 | 55.93 | 78.09 | 128.78 | 127.75 | 158.89 | 2.72 | 0.629 | 0.248 |
| CHB-MIT | 59.92 | 58.61 | 78.00 | 101.21 | 95.87 | 130.41 | 2.00 | 0.750 | 0.875 |
| ds005292 | 72.20 | 70.87 | 80.07 | 116.45 | 114.82 | 132.79 | 2.03 | 2.400 | 2.656 |
| Mean of four sources | 63.11 | | | 122.43 | | | 2.44 | | |

| | Amortized latency per window (ms), EEG-Jev / AR | | | | |
|---|---|---|---|---|---|
| Batch size | Lee2019 | Sleep-EDF-SC | CHB-MIT | ds005292 | Mean of four sources |
| 1 | 61.16 / 143.30 | 59.16 / 128.78 | 59.92 / 101.21 | 72.20 / 116.45 | 63.11 / 122.43 |
| 2 | 40.30 / 80.86 | 31.81 / 68.94 | 30.95 / 53.13 | 50.00 / 72.84 | 38.27 / 68.94 |
| 4 | 32.09 / 53.61 | 21.39 / 39.49 | 21.44 / 31.49 | 39.81 / 51.47 | 28.68 / 44.01 |
| 8 | 32.37 / 42.40 | 19.13 / 26.93 | 20.42 / 25.05 | 41.17 / 47.71 | 28.27 / 35.52 |
| 16 | 33.77 / 38.63 | 19.24 / 21.10 | 20.71 / 22.35 | 42.66 / 46.38 | 29.09 / 32.12 |
| 32 | 34.46 / 36.14 | 19.41 / 19.10 | 20.72 / 21.48 | 43.41 / 46.52 | 29.50 / 30.81 |

**Timing procedure.** Both paths run on one NVIDIA RTX PRO 6000 Blackwell GPU with PyTorch 2.8.0, 32-bit floating-point weights and arithmetic, and one CPU thread; the optimized configuration adds fused kernels for the Gated DeltaNet layers of the language model, set to IEEE arithmetic. The two paths are timed alternately on the same windows after a warm-up call, at batch sizes from 1 to 32 with batches formed within a source, and the GPU is synchronized before the start and at the end of every timed call. Every window is timed three times at batch sizes up to 8 and thirteen times at 16 and 32. An AR answer is valid only if the complete text is a listed candidate identifier or name, “yes” or “no”, or a number from 0 to 10.

**Inputs and answers.** The windows have 62, 2, 24 and 63 channels (Lee2019, Sleep-EDF-SC, CHB-MIT, ds005292). The input row of EEG-Jev has 345, 196, 213 and 464 positions and the prefill of the AR comparator 333, 184, 197 and 454; the AR row is shorter mainly because its instruction is shorter than the system prompt of EEG-Jev. The AR answers have a mean length of 3.00, 2.72, 2.00 and 2.03 tokens including the end-of-sequence token. In a batch, rows that have finished are carried along until the last row ends, so the longest answer determines the time of the batch.

**Adaptation of the comparator.** Without adaptation the EEG-Jev checkpoint does not produce usable text: on a diagnostic panel of 32 development windows, disjoint from the timing panel, only 2 answers were valid and 24 reached the 32-token limit. After the 50 LoRA updates (800 training questions, 200 per application; AdamW, learning rate $2 \times 10^{-5}$) all 32 were valid and ended with the end-of-sequence token. The number of updates was chosen on this small panel.

## E Results for all sources

Table A4 gives the raw-output test results of EEG-Jev, B1 and B2 for each of the 35 sources in the main metric of its application. Paired intervals come from 2,000 resamples of the test subjects or patients within a source and are given for the 26 sources with at least five such independent units; for the other nine the interval is omitted, a rule fixed before the test evaluation. The intervals describe the sampling of the test population. They do not include variation between training seeds: EEG-Jev has one training run, and each baseline value is the mean of two seeds, whose individual results are in the result files. We report no application-level averages, because the sources of an application differ in size, protocol and number of classes and metrics of different applications are not commensurable.

**Pain reference.** On ds005292, predicting the training-set mean gives an MAE of 1.771, compared with 1.700 for EEG-Jev, 1.701 for B1 and 1.717 for B2. The paired intervals between EEG-Jev and the learned baselines include zero.

**NeuralBench best.** The NeuralBench values of Table 2 are the highest mean test BA among the entries that the official NeuralBench-EEG v1.0 results table lists for each core task[15,16] (results file at commit 7aaccf3 of the NeuralBench repository, package release 0.4.0, unchanged on 12 November 2026), given as mean and standard deviation over three seeds: REVE-Base on Stieger2021, CBraMod on Sleep-EDF-SC and BDTCN on CHB-MIT. They share with our evaluation the partition files, the windows, the labels and the metric. They differ in the model and its training (one model per task, trained or fine-tuned end to end with checkpoint selection on development BA), in the signal recipe of the sleep and seizure entries, and in the number of seeds.

**Complete metric results.** The results/ archive on the project website holds every applicable metric that was computed, for every source, model, training seed and output stage (raw, prior-corrected, temperature-scaled): metrics-long.csv (one record per metric value, with threshold, direction, sample size, number of independent units and interval status), metrics-wide-by-source.csv, metrics-by-seed.csv, paired-differences.csv (differences, intervals and the reason for each missing interval) and data-dictionary.md (metric definitions). The values behind the tables of this paper and the inference benchmark are distributed in the same archive. No per-window test data is released.

**Table A4. Test results per source on the display metric, raw outputs: BA in percent (motor imagery, sleep staging); BA at raw $p(\text{yes}) \geq 0.5$ in percent and AUROC (seizure detection); MAE on the 0–10 scale (pain rating).** Units: independent test units (subjects; patients for seizure). B1, B2: mean of seeds 7 and 8. $\Delta$: EEG-Jev minus baseline (MAE: baseline minus EEG-Jev), positive when EEG-Jev is better, in the unit of the row. Brackets: paired 95% interval from 2,000 resamples of the test subjects or patients within the source. It describes sampling of the test population only; variation between training seeds is not included (EEG-Jev has one seed; per-seed baseline values are in the result files). Intervals are omitted when the source has fewer than five independent test units (9 sources), a rule fixed before the test evaluation. Bold marks the best displayed value among EEG-Jev, B1 and B2 in each row; displayed ties are bold together. $*$: core source. Other computed metrics (accuracy, macro-F1, AUROC, average precision, sensitivity and specificity, NLL, Brier score, ECE, CRPS, coverage; every seed and output stage) are in results/metrics-long.csv; results/README.md indexes the result files.

| Source | Units | EEG-Jev | B1 | B2 | $\Delta$ vs B1 [95% interval] | $\Delta$ vs B2 [95% interval] |
|---|---|---|---|---|---|---|
| *Motor imagery, cross-subject, K=4; BA (%)* | | | | | | |
| EEGMMIDB | 22 | 59.09 | 58.32 | **66.41** | +0.77 [−1.31, +2.98] | −7.32 [−10.33, −4.35] |
| Stieger2021* | 13 | 54.08 | 61.95 | **65.19** | −7.87 [−10.66, −5.52] | −11.11 [−13.91, −8.39] |
| BCI-IV-2a | 2 | 44.18 | 33.55 | **45.14** | +10.63 | −0.95 |
| *Motor imagery, cross-subject, K=2; BA (%)* | | | | | | |
| Dreyer2023 | 21 | 76.35 | 77.35 | **80.35** | −1.00 [−4.26, +2.10] | −4.00 [−6.88, −1.21] |
| Lee2019 | 11 | 80.86 | 73.57 | **83.86** | +7.30 [+4.84, +9.98] | −3.00 [−4.77, −1.16] |
| Cho2017 | 11 | 72.23 | 71.41 | **75.95** | +0.82 [−1.16, +3.20] | −3.73 [−5.70, −1.75] |
| Liu2022 | 10 | **51.75** | 49.13 | 50.63 | +2.63 [−5.13, +9.63] | +1.13 [−7.88, +8.88] |
| Shin2017A | 6 | 60.28 | 59.31 | **65.56** | +0.97 [−4.86, +7.78] | −5.28 [−11.53, +0.83] |
| Faller2012 | 3 | 65.67 | 62.92 | **66.17** | +2.75 | −0.50 |
| Scherer2012 | 3 | 58.33 | 56.15 | **58.65** | +2.19 | −0.31 |
| BNCI2014-004 | 2 | **77.50** | 71.70 | 74.41 | +5.80 | +3.09 |
| GrosseWentrup2009 | 2 | 55.00 | 49.17 | **58.50** | +5.83 | −3.50 |
| *Motor imagery, within-subject, K=2–3; BA (%)* | | | | | | |
| Barachant2012 | 8 | 34.38 | **37.50** | 32.81 | −3.13 [−13.33, +9.72] | +1.56 [−7.28, +14.41] |
| Dornhege2004 | 5 | **66.07** | 58.04 | 65.63 | +8.04 [−1.11, +13.31] | +0.45 [−6.17, +9.74] |
| Zhou2016 | 4 | **70.50** | 65.67 | 64.83 | +4.83 | +5.67 |
| BEETL-A | 3 | **41.44** | 35.67 | 36.28 | +5.78 | +5.17 |
| BEETL-B | 2 | **52.00** | 38.92 | 37.83 | +13.08 | +14.17 |

| Source | Units | EEG-Jev | B1 | B2 | *Δ* vs B1 [95% interval] | *Δ* vs B2 [95% interval] |
|---|---|---|---|---|---|---|
| *Sleep staging, K=5; BA (%)* | | | | | | |
| CinC2018 | 199 | 74.17 | 75.74 | **75.83** | −1.56 [−2.15, −0.96] | −1.66 [−2.23, −1.08] |
| HMC | 31 | **74.87** | 72.16 | 72.51 | +2.71 [+0.87, +4.74] | +2.36 [+0.52, +4.52] |
| CAP | 22 | **70.32** | 67.84 | 68.29 | +2.48 [+0.23, +5.16] | +2.03 [−0.35, +4.77] |
| Sleep-EDF-SC* | 16 | 70.45 | 68.17 | **70.48** | +2.29 [+0.46, +3.94] | −0.03 [−1.98, +1.93] |
| DOD-O | 12 | 65.14 | **67.54** | 67.41 | −2.40 [−5.22, +0.68] | −2.27 [−5.85, +1.46] |
| DOD-H | 6 | 71.32 | **71.65** | 70.21 | −0.33 [−4.38, +3.93] | +1.11 [−1.53, +4.12] |
| Sleep-EDF-ST | 5 | **76.73** | 69.53 | 71.55 | +7.20 [+1.14, +13.99] | +5.18 [+0.49, +9.79] |
| *Seizure detection; BA at 0.5 (%)* | | | | | | |
| CHB-MIT* | 5 | **72.11** | 68.81 | 68.65 | +3.30 [−0.46, +6.53] | +3.46 [−0.15, +6.73] |
| Siena | 4 | **74.53** | 74.41 | 70.34 | +0.13 | +4.20 |
| *Seizure detection; AUROC* | | | | | | |
| CHB-MIT* | 5 | 0.825 | 0.891 | **0.929** | −0.066 [−0.141, −0.005] | −0.105 [−0.185, −0.047] |
| Siena | 4 | 0.804 | **0.865** | 0.838 | −0.061 | −0.034 |
| *Pain rating; MAE* | | | | | | |
| ds005292* | 26 | **1.700** | 1.701 | 1.717 | +0.001 [−0.131, +0.136] | +0.017 [−0.080, +0.113] |
| ds005280 | 20 | **1.609** | 1.613 | 1.636 | +0.004 [−0.271, +0.273] | +0.027 [−0.247, +0.281] |
| ds005293 | 19 | 1.958 | 1.953 | **1.914** | −0.006 [−0.089, +0.069] | −0.044 [−0.159, +0.086] |
| ds005291 | 13 | **1.402** | 1.581 | 1.498 | +0.178 [+0.057, +0.283] | +0.096 [−0.021, +0.196] |
| ds005289 | 8 | **1.303** | 1.329 | 1.330 | +0.026 [−0.135, +0.205] | +0.028 [−0.126, +0.201] |
| ds005285 | 6 | **2.225** | 2.331 | 2.245 | +0.106 [−0.055, +0.333] | +0.020 [−0.364, +0.282] |
| ds005286 | 6 | 2.037 | **1.951** | 2.051 | −0.086 [−0.336, +0.164] | +0.014 [−0.425, +0.488] |
| ds005473 | 6 | 1.259 | 1.346 | **1.188** | +0.086 [−0.083, +0.221] | −0.071 [−0.205, +0.075] |
| ds005284 | 5 | 1.690 | **1.602** | 1.634 | −0.088 [−0.476, +0.308] | −0.056 [−0.519, +0.401] |